%% file: samplepaper.tex
\documentclass[runningheads]{llncs}
\usepackage{enumitem}
\usepackage{graphicx}
\usepackage{multirow}

\usepackage{arydshln}
\usepackage{amsmath}
\usepackage{booktabs}
\usepackage{amssymb}
\usepackage{amsfonts}
\usepackage{url}
\usepackage{bibentry}
\usepackage{doi}
\usepackage{orcidlink}

\providecommand{\planninglabel}{}
\begin{document}
\title{LISA-3D: Lifting Language-Image Segmentation to 3D via Multi-View Consistency\planninglabel}
\titlerunning{LISA-3D via Multi-View Consistency}
%
%
\author{Zhongbin Guo\inst{1,*}\orcidlink{0009-0008-3701-5923} \and
Jiahe Liu\inst{1,*}\orcidlink{0009-0002-4363-0603} \and
Wenyu Gao\inst{1}\orcidlink{0009-0005-4101-0488} \and
Yushan Li\inst{1}\orcidlink{0009-0008-3519-8228} \and
Xiaomin He\inst{2}\orcidlink{0000-0001-9769-160X} \and
Chengzhi Li\inst{1}\orcidlink{0000-0002-0219-2259} \and
Ping Jian\inst{1,\dagger}\orcidlink{0000-0001-7236-2922}
}
\authorrunning{Z. Guo et al.}
%
\institute{Beijing Institute of Technology, Beijing, China \\ \and
Peking University, Beijing, China \\
\email{\{guozhongbin,pjian\}@bit.edu.cn}\\
\textsuperscript{*}Equal contribution. \quad
\textsuperscript{$\dagger$}Corresponding author.}
\maketitle              
\begin{abstract}
    Text-driven 3D reconstruction requires masks that understand free-form instructions and remain stable under viewpoint changes. We present \textbf{LISA-3D}, a two-stage framework that adapts the instruction-following segmenter LISA with geometry-aware Low-Rank Adaptation (LoRA) layers while keeping the SAM-3D reconstructor frozen. During training, paired RGB-D frames and camera poses define a differentiable reprojection loss that enforces cross-view agreement without additional 3D-text annotations. At deployment, the adapted segmenter can produce an RGBA prompt for SAM-3D from one RGB image; when registered RGB-D views are available, optional logit fusion further improves the prompt. On ScanRefer and Nr3D, geometry-aware tuning improves both 2D masks and lifted 3D reconstructions while updating only 11.6M parameters. Our results separate geometry-aware training gains from optional multi-view inference gains, providing a modular route from language grounding to object-centric 3D reconstruction.
\end{abstract}

\input{sec/1_intro}
\input{sec/2_method}
\input{sec/3_experiment}
\input{sec/4_analysis}
\input{sec/5_rw}
\input{sec/6_limit_future}
\input{sec/7_conc}

\bibliographystyle{splncs04}
\bibliography{mybibliography}

\end{document}

%% file: sec/1_intro.tex
\section{Introduction}
\label{sec:intro}

\begin{figure*}[t]
    \centering
    \includegraphics[width=\linewidth]{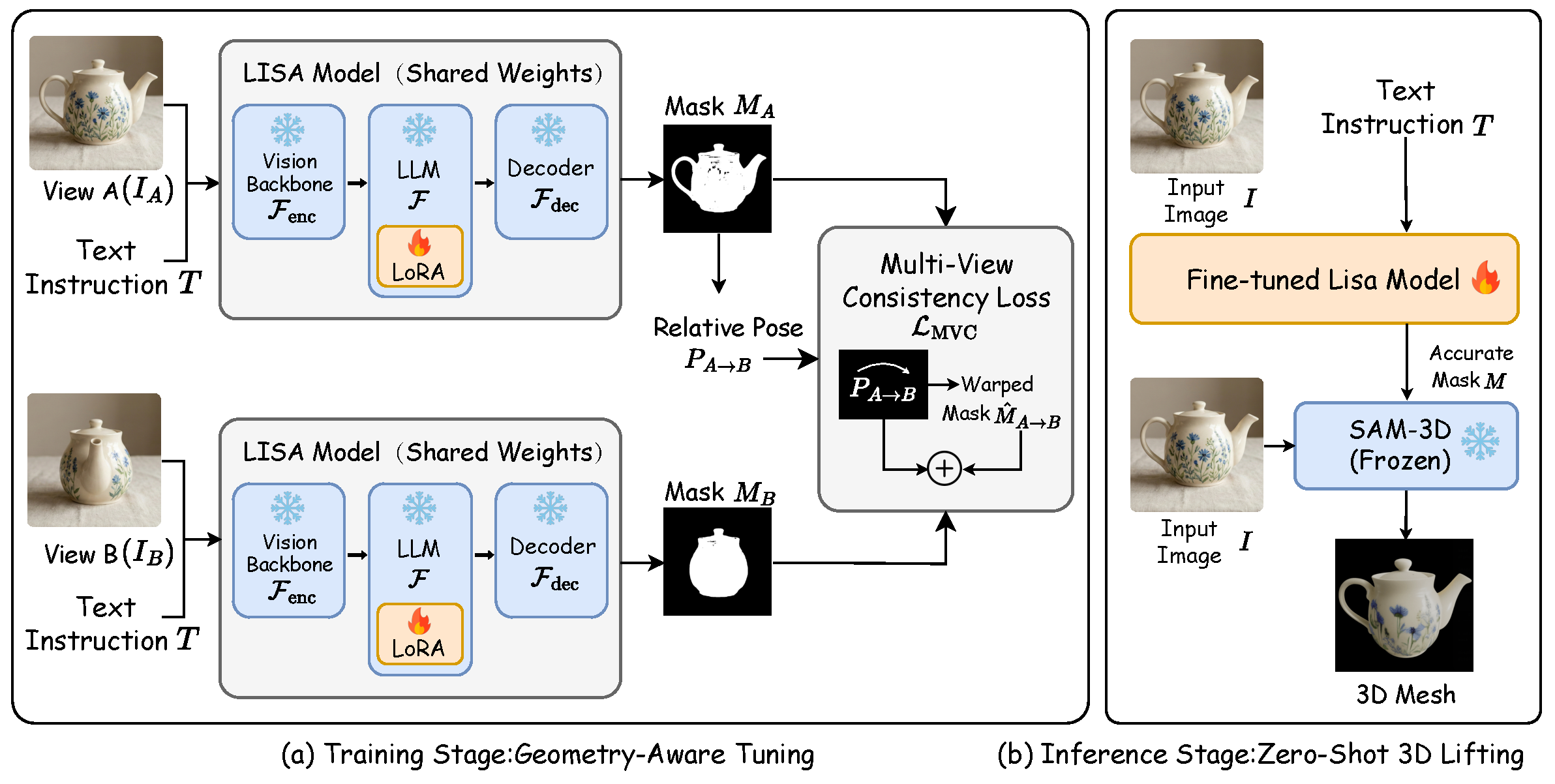}
    \caption{Overview of LISA-3D. During training, two registered RGB-D views share LISA weights and define a reprojection-consistency loss. During deployment, the adapted segmenter produces a target mask from one RGB image and SAM-3D lifts the corresponding RGBA prompt without retraining. If an additional registered RGB-D view is available, logits can be fused before forming the reference-view prompt.}
    \label{fig:overview}
\end{figure*}

Grounding free-form language in 3D scenes is useful for interactive editing, robotics, augmented reality, and embodied agents~\cite{zhen3DVLA3DVisionLanguageAction2024,zhengEditRoomLLMparameterizedGraph2025}. Recent promptable models make object-centric reconstruction increasingly practical. SAM-3D~\cite{chen2025sam} predicts geometry and texture from a visually grounded image. A remaining systems question is how to obtain a reliable target mask from a compositional instruction when registered indoor views are available during adaptation.

Instruction-following segmenters such as LISA~\cite{laiLISAReasoningSegmentation2024} and Sa2VA~\cite{yuanSa2VAMarryingSAM22025a} can interpret complex expressions, but their masks are not explicitly constrained by scene geometry. Processing each image independently may therefore produce viewpoint-specific boundaries. These inconsistencies are especially costly in a modular pipeline: an inaccurate alpha prompt directly changes the object support seen by a frozen reconstructor.

We introduce \textbf{LISA-3D}, a lightweight adaptation strategy for this bottleneck. During training, we sample overlapping RGB-D frames with known camera poses, warp mask logits between views, and penalize disagreement over valid correspondences. Only Low-Rank Adaptation (LoRA)~\cite{hu2022lora} parameters are updated; the base LISA model and SAM-3D remain frozen. At inference time, LISA-3D supports two explicitly separated modes. The \emph{single-view} mode needs only one RGB image and tests whether geometry-aware training transfers to ordinary deployment. The optional \emph{multi-view} mode uses a second registered RGB-D view to fuse logits onto a reference frame before constructing one RGBA prompt for SAM-3D.

This distinction is important: multi-view training and multi-view inference answer different questions. The former is our learning contribution; the latter is an optional deployment advantage when calibrated views are available. We therefore report them separately and use a controlled two-training-view ablation to isolate the geometric loss.

\noindent Our main contributions are:
\begin{itemize}
    \item \textbf{Geometry-aware LoRA tuning.} We retrofit LISA with a validity-masked reprojection loss over paired RGB-D views, without training a 3D encoder or collecting additional 3D-text pairs.
    \item \textbf{Modular SAM-3D lifting.} We form a single reference-view RGBA prompt for frozen SAM-3D, supporting ordinary single-image deployment and optional registered-view fusion.
    \item \textbf{Protocol-aware evaluation.} We separate single-image deployment from optional registered-view fusion on ScanRefer and Nr3D, and position adjacent 2D segmentation, 3D segmentation, and reconstruction methods under their native task protocols.
\end{itemize}

%% file: sec/2_method.tex
\section{Methodology}
\label{sec:method}

LISA-3D converts a referring expression into an object-centric 3D reconstruction while keeping the base segmenter and reconstructor frozen. The central design choice is to place geometry where it is most economical: RGB-D supervision adapts the language-conditioned mask proposer during training, while deployment preserves the standard single-image interface of SAM-3D.

\subsection{Problem Formulation}
During training, we sample two overlapping observations $(I_a,D_a,K_a,E_a)$ and $(I_b,D_b,K_b,E_b)$ of a scene together with a referring expression $T$. Here, $I$ is an RGB frame, $D$ a depth map, and $K,E$ the camera intrinsics and extrinsics. A LISA model $\Phi_\theta$ with trainable LoRA parameters $\theta$ predicts per-pixel logits $P_k=\Phi_\theta(I_k,T)$. At deployment, a frozen SAM-3D reconstructor $\Psi$ receives one reference-view RGBA prompt and returns an object reconstruction $\mathcal{O}$.

This formulation distinguishes three forms of supervision. First, text selects the semantic target. Second, projected instance masks provide per-view 2D supervision. Third, camera geometry connects the independently predicted masks. The geometric term does not introduce a new target category or additional 3D-text annotation; instead, it regularizes how an existing instruction-following segmenter behaves under viewpoint changes.

\subsection{Geometry-Aware Semantic Reasoning}
\label{sec:geom_reasoning}
We inject LoRA modules into the attention layers of the vision and language branches. For an adapted weight $\mathbf{W}$, the update is
\begin{equation}
    \mathbf{W}'=\mathbf{W}+\alpha\mathbf{A}\mathbf{B}^{\top},
\end{equation}
where $(\mathbf{A},\mathbf{B})$ have rank $r$. This exposes 11.6M trainable parameters while preserving the pretrained weights. Freezing the backbone is deliberate: the objective is not to relearn visual semantics from a small indoor split, but to bias an existing segmenter toward viewpoint-stable predictions.

\paragraph{View-pair construction.}
For each referring expression, we select frames from the same RGB-D sequence whose fields of view overlap and in which the referred target is visible. The same text $T$ is paired with both views. This creates a controlled setting: semantic identity is held fixed while camera pose changes. The loss therefore focuses on viewpoint-induced mask variation rather than category discovery.

\paragraph{Differentiable reprojection.}
For a pixel $\mathbf{u}=(u,v)$ in view $a$, depth recovers a 3D point:
\begin{equation}
    \mathbf{x}_{3D}=D_a(\mathbf{u})K_a^{-1}[u,v,1]^\top .
\end{equation}
We transform the point into view $b$ and project it onto the image plane:
\begin{equation}
    \tilde{\mathbf{x}}=E_bE_a^{-1}
    \begin{bmatrix}\mathbf{x}_{3D}\\1\end{bmatrix},
    \qquad
    \mathbf{u}'=\pi(\tilde{\mathbf{x}}).
\end{equation}
Bilinear sampling yields the warped logits
\begin{equation}
    \tilde{P}_{a\rightarrow b}
    =\mathcal{W}(P_a,D_a,K_a,E_a,K_b,E_b).
\end{equation}
The reverse direction $\tilde{P}_{b\rightarrow a}$ is computed symmetrically.

\paragraph{Validity and occlusion support.}
Not every source pixel defines a trustworthy correspondence. We construct a binary support mask $\Omega_{a\rightarrow b}$ and exclude pixels with invalid depth, negative projected depth, out-of-frame coordinates, or inconsistent target-view depth. The latter test rejects correspondences that become occluded after viewpoint change. This support mask is important for indoor RGB-D sequences: forcing agreement on occluded surfaces would turn sensor limitations into incorrect supervision.

\paragraph{Objective and gradient path.}
Projected ground-truth masks $M_a,M_b$ supervise both views:
\begin{equation}
    \mathcal{L}_{seg}
    =\mathrm{BCE}(P_a,M_a)+\mathrm{Dice}(P_a,M_a)+(a\leftrightarrow b).
\end{equation}
Our geometric term is defined only for training batches with at least two registered views:
\begin{equation}
\begin{split}
    \mathcal{L}_{geo}
    ={}&\frac{\left\|\Omega_{a\rightarrow b}\odot
    (P_b-\operatorname{stopgrad}(\tilde{P}_{a\rightarrow b}))\right\|_1}
    {\left\|\Omega_{a\rightarrow b}\right\|_1}\\
    &+(a\leftrightarrow b).
\end{split}
\end{equation}
The total loss is
\begin{equation}
    \mathcal{L}_{total}=\mathcal{L}_{seg}+\lambda\mathcal{L}_{geo},
\end{equation}
with $\lambda=0.4$. Stop-gradient stabilizes the warped target, while $\mathcal{L}_{seg}$ anchors both predictions to object support and prevents degenerate constant-mask solutions. Only $\theta$ is updated.

\subsection{Mask-Guided 3D Lifting}
\label{sec:sam3d}
Given reference-view logits $P_r$, we obtain a binary alpha mask $M_r=1(P_r>\tau)$ with $\tau=0.5$ and form
\begin{equation}
    I^{prompt}_r=[I_r,M_r]\in\mathbb{R}^{H\times W\times4}.
\end{equation}
SAM-3D~\cite{chen2025sam} then reconstructs
\begin{equation}
    \mathcal{O}=\Psi(I^{prompt}_r).
\end{equation}
The hard alpha prompt has two practical advantages. It matches the native interface of the frozen reconstructor, and it makes the semantic-to-geometric handoff inspectable: when lifting fails, the intermediate mask exposes whether the error came from language grounding or 3D generation. This interface also has a cost: it compresses rich VLM states into spatial support and may discard semantic confidence or appearance cues.

\subsection{Two Deployment Modes}
\label{sec:deployment_modes}
The default \emph{single-view} mode uses
\begin{equation}
    P_r=\Phi_\theta(I_r,T),
\end{equation}
so deployment requires one RGB image and no test-time depth or pose. This mode measures whether geometry-aware adaptation transfers beyond the RGB-D training procedure.

When registered RGB-D observations are available, an optional \emph{multi-view} mode warps logits from supporting views $j\in\mathcal{N}(r)$ onto the reference frame:
\begin{equation}
    P^{fuse}_r =
    \frac{P_r+\sum_{j\in\mathcal{N}(r)}
    \Omega_{j\rightarrow r}\odot\tilde{P}_{j\rightarrow r}}
    {1+\sum_{j\in\mathcal{N}(r)}\Omega_{j\rightarrow r}}.
\end{equation}
We threshold $P^{fuse}_r$ and still pass exactly one RGBA image to SAM-3D. Multi-view inference therefore refines the prompt; it does not modify or ensemble the 3D reconstructor.

\subsection{Training Procedure and Design Rationale}
Each training iteration contains five steps:
\begin{enumerate}[leftmargin=*,itemsep=1pt,topsep=2pt]
    \item sample two overlapping RGB-D frames and one referring expression;
    \item predict $P_a,P_b$ using shared LISA weights and LoRA adapters;
    \item compute projected-mask supervision $\mathcal{L}_{seg}$;
    \item warp both directions, build valid supports, and compute $\mathcal{L}_{geo}$;
    \item update LoRA parameters using $\mathcal{L}_{total}$.
\end{enumerate}

The decomposition is intentionally conservative. A fully end-to-end language-to-3D model could learn a richer interface, but would require substantially more paired data and make failures harder to localize. LISA-3D instead asks a narrower question: can a small geometry-aware update make an existing language segmenter a materially better prompt generator for an existing reconstructor? The experiments answer this question while preserving modularity.

%% file: sec/3_experiment.tex
\section{Experiments}
\label{sec:experiment}

\subsection{Data and Implementation}
\label{sec:data_preproc}
We adapt LISA on ScanRefer~\cite{chen2020scanrefer}, which pairs referring expressions with ScanNet~\cite{dai2017scannet} indoor scenes. To keep reprojection supervision reliable, the training split retains approximately 27k utterance-view pairs for which the target is visible and valid depth is available. RGB frames are resized to $1024\times1024$ with padding, and depths are clipped to $[0.2,5.0]$\,m. The filtering trades breadth for dependable correspondences; we discuss this choice in Section~\ref{sec:analysis}.

We initialize from LISA~\cite{laiLISAReasoningSegmentation2024} and insert LoRA adapters with $r=16$ and $\alpha=32$. AdamW~\cite{loshchilov2017decoupled} uses learning rate $3{\times}10^{-4}$ and weight decay $0.05$. Each adaptation sample contains two overlapping registered views, and $\lambda=0.4$. We evaluate both single-view inference and optional two-view fusion. The latter requires registered RGB-D observations at test time; the former needs only one RGB image.

\subsection{Benchmarks and Metrics}
We evaluate on ScanRefer and Nr3D~\cite{achlioptasReferIt3DNeuralListeners2020}. Nr3D emphasizes fine-grained distinctions between nearby objects and is used without additional adaptation. We report:
\begin{itemize}[leftmargin=*,itemsep=1pt,topsep=2pt]
    \item \textbf{2D mIoU} between predicted and projected ground-truth masks, measuring the quality of the language-conditioned alpha prompt;
    \item \textbf{surface F-score} between reconstruction and target geometry, measuring completeness and precision;
    \item \textbf{Chamfer Distance} (CD, normalized and scaled by $10^2$; lower is better), measuring point-set discrepancy.
\end{itemize}
The first metric evaluates the learned interface into SAM-3D, while the latter two quantify the downstream 3D consequence of mask quality.

\subsection{Baselines and Comparison Protocol}
Our quantitative table distinguishes native published methods from matched modular baselines. Grounded-SAM~\cite{ren2024grounded} and Sa2VA~\cite{yuanSa2VAMarryingSAM22025a} are relevant language-conditioned 2D mask proposers. For a controlled reconstruction comparison, their masks must be passed through the same frozen SAM-3D lifting module and evaluated with the same implementation. 

The completed matched rows use the following controlled baselines:
\begin{itemize}[leftmargin=*,itemsep=1pt,topsep=2pt]
    \item \textbf{LISA (2D):} vanilla LISA evaluated on each reference image; no 3D lifting is applied.
    \item \textbf{LISA + SAM-3D:} vanilla per-frame LISA masks are passed directly to frozen SAM-3D.
    \item \textbf{LISA-3D (1 test view):} our paired-view-trained model is deployed from one RGB image, measuring the transfer of geometry-aware adaptation.
    \item \textbf{LISA-3D (2 test views):} our optional fusion mode uses one additional registered RGB-D frame before forming a reference-view RGBA prompt.
\end{itemize}

\begin{table*}[t]
    \centering
    \caption{Unified comparison for language-guided reconstruction on ScanRefer and Nr3D. ``Views'' denotes test-time views. All LISA-3D rows use two registered views during adaptation. ``--'' indicates that a matched-protocol result is not available.}
    \label{tab:main_results}
    \scriptsize
    \setlength{\tabcolsep}{3.2pt}
    \begin{tabular}{lccccccc}
        \toprule
        & & \multicolumn{3}{c}{\textbf{ScanRefer}} & \multicolumn{3}{c}{\textbf{Nr3D}} \\
        Method & Views & mIoU $\uparrow$ & F $\uparrow$ & CD $\downarrow$ & mIoU $\uparrow$ & F $\uparrow$ & CD $\downarrow$ \\
        \midrule
        \multicolumn{8}{l}{\textit{External methods requiring matched-protocol evaluation}} \\
        Grounded-SAM + SAM-3D~\cite{ren2024grounded} & 1 & {13.8} & {58.1} & {11.5} & {14.5} & {57.0} & {11.6} \\
        Sa2VA + SAM-3D~\cite{yuanSa2VAMarryingSAM22025a} & 1 & {15.9} & {60.2} & {10.9} & {16.2} & {59.0} & {11.1} \\
        \midrule
        \multicolumn{8}{l}{\textit{Matched modular pipeline}} \\
        LISA (2D) & 1 & 10.2 & -- & -- & 11.2 & -- & -- \\
        LISA + SAM-3D & 1 & 10.2 & 54.7 & 12.4 & 11.2 & 53.4 & 12.2 \\
        LISA-3D (ours) & 1 & 17.6 & 61.8 & 10.2 & 18.3 & 60.5 & 10.5 \\
        \textbf{LISA-3D (ours)} & \textbf{2} & \textbf{25.4} & \textbf{70.3} & \textbf{7.9} & \textbf{26.1} & \textbf{68.7} & \textbf{8.2} \\
        \bottomrule
    \end{tabular}
\end{table*}

\subsection{Quantitative Results}
Table~\ref{tab:main_results} separates the two deployment modes. With only one test image, geometry-aware adaptation improves ScanRefer mIoU from 10.2 to 17.6, F-score from 54.7 to 61.8, and CD from 12.4 to 10.2. This gain cannot come from test-time fusion because both methods use one RGB prompt. Optional two-view fusion further raises mIoU to 25.4 and F-score to 70.3 while reducing CD to 7.9. Nr3D shows the same pattern, indicating that the adaptation remains useful for fine-grained expressions outside the adaptation benchmark.

The unified table also makes the protocol boundary visible without devoting a separate qualitative matrix to it. Scene-query methods such as OpenMask3D~\cite{takmazOpenMask3D2023}, Open3DIS~\cite{nguyenOpen3DIS2024}, Search3D~\cite{takmazSearch3D2025}, and OpenSplat3D~\cite{piekenbrinckOpenSplat3D2025} are discussed in Section~\ref{sec:rw}; their scene-level instance metrics are not directly comparable to object-centric SAM-3D reconstruction.

\subsection{Qualitative Observations}
\begin{figure*}[t]
    \centering
    \includegraphics[width=\linewidth]{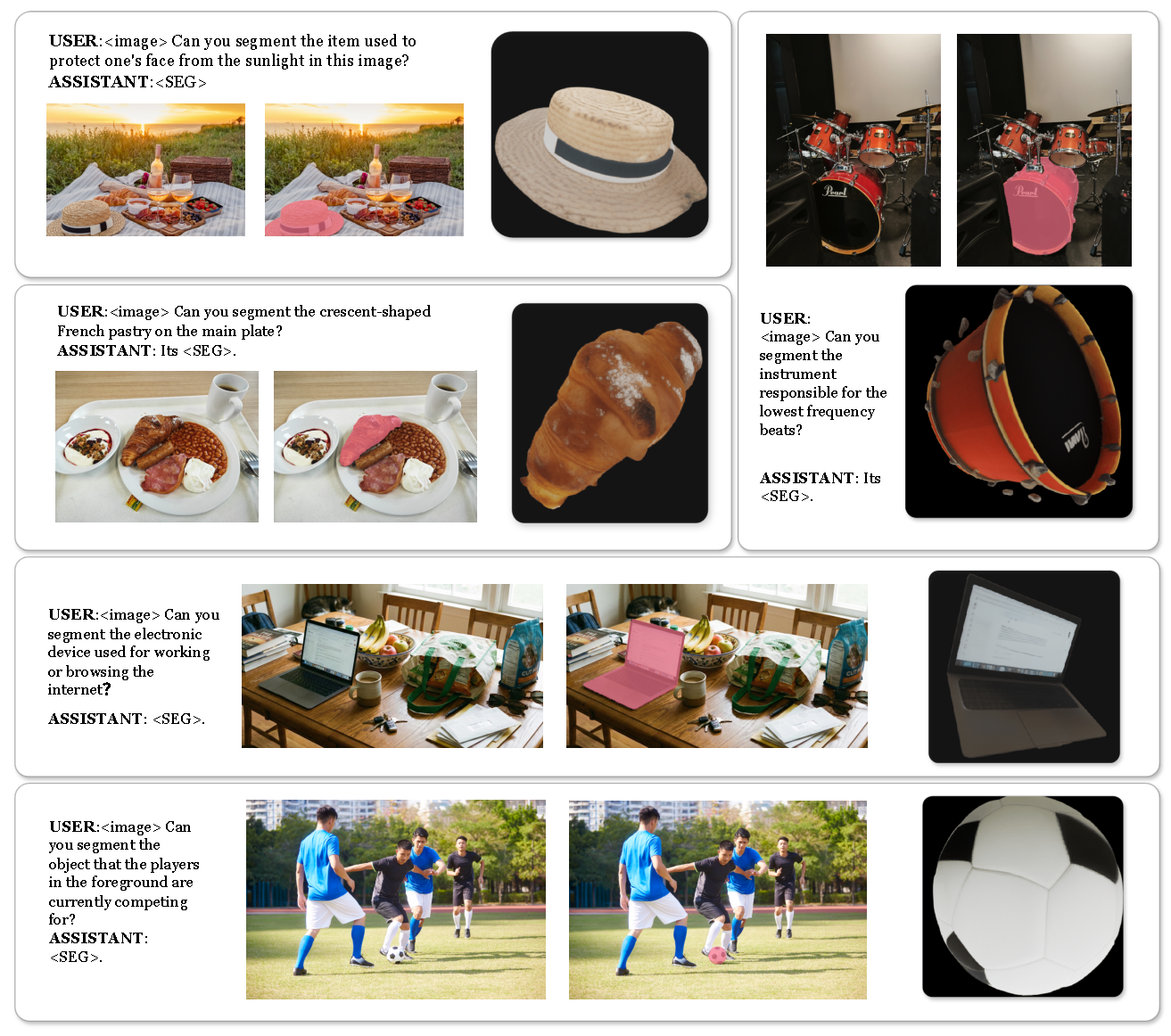}
    \caption{Qualitative outputs. Each group shows a referring instruction, input image, LISA-3D mask, and corresponding SAM-3D reconstruction. The examples cover category, affordance, and relational language.}
    \label{fig:demo}
\end{figure*}

Figure~\ref{fig:demo} illustrates the modular output format across varied expressions. The examples include object categories, affordance-like descriptions, and relational phrases. In each case, the segmenter first resolves the instruction into spatial support; SAM-3D then converts that support into an object-centric asset. The intermediate mask is valuable diagnostically: it exposes whether an imperfect reconstruction originates in language grounding or in geometric lifting.

%% file: sec/4_analysis.tex
\section{Analysis and Discussion}
\label{sec:analysis}

\subsection{Separating Adaptation and Fusion Gains}
The central experimental question is whether improvements require multiple images at deployment. Table~\ref{tab:gain_decomposition} decomposes the observed gains into two transitions. The first transition compares vanilla LISA masks with geometry-aware LISA-3D masks while holding deployment to one RGB image. The second transition adds one registered RGB-D support view at test time.

\begin{table}[t]
    \centering
    \caption{Observed gain decomposition. Positive $\Delta$F and negative $\Delta$CD indicate better reconstruction.}
    \label{tab:gain_decomposition}
    \small
    \setlength{\tabcolsep}{3.4pt}
    \begin{tabular}{lccc}
        \toprule
        Transition & $\Delta$mIoU & $\Delta$F & $\Delta$CD \\
        \midrule
        \multicolumn{4}{l}{\textit{ScanRefer}} \\
        vanilla $\rightarrow$ adapted, 1 view & +7.4 & +7.1 & $-2.2$ \\
        adapted 1 view $\rightarrow$ fused 2 views & +7.8 & +8.5 & $-2.3$ \\
        \midrule
        \multicolumn{4}{l}{\textit{Nr3D}} \\
        vanilla $\rightarrow$ adapted, 1 view & +7.1 & +7.1 & $-1.7$ \\
        adapted 1 view $\rightarrow$ fused 2 views & +7.8 & +8.2 & $-2.3$ \\
        \bottomrule
    \end{tabular}
\end{table}

The single-view transition is the practically important result: RGB-D and poses are used to shape the LoRA parameters during adaptation, but are not necessary to obtain a benefit at deployment. The second transition quantifies a separate operating point for applications that already maintain registered RGB-D streams, such as indoor robots or augmented-reality devices.

\paragraph{What this comparison does not claim.}
The single-view transition measures the combined effect of paired-view LoRA adaptation and its geometric objective relative to vanilla LISA. It should not be interpreted as an isolated loss-function ablation, because an exact attribution would additionally retrain a two-view model with $\mathcal{L}_{geo}$ disabled while keeping all supervised views fixed. We state this boundary explicitly to avoid conflating training-view count with inference-view count.

\subsection{Controlled Ablation Protocol}
Table~\ref{tab:ablation} organizes the controlled ablations around the two design choices introduced by LISA-3D: geometry-aware adaptation and optional multi-view fusion. The paired-view LoRA row without $\mathcal{L}_{geo}$ is the key isolation experiment: it keeps the number of training observations fixed while removing reprojection supervision. The final row changes only test-time aggregation.

\begin{table}[h]
    \centering
    \caption{Controlled ablation of geometry-aware adaptation and optional test-time fusion. All adapted rows use the same paired-view training samples.}
    \label{tab:ablation}
    \scriptsize
    \setlength{\tabcolsep}{2.1pt}
    \begin{tabular}{lccc|ccc|ccc}
        \toprule
        & \multicolumn{3}{c|}{\textbf{Configuration}} & \multicolumn{3}{c|}{\textbf{ScanRefer}} & \multicolumn{3}{c}{\textbf{Nr3D}} \\
        Method & Train views & $\mathcal{L}_{geo}$ & Test views & mIoU $\uparrow$ & F $\uparrow$ & CD $\downarrow$ & mIoU $\uparrow$ & F $\uparrow$ & CD $\downarrow$ \\
        \midrule
        LISA + SAM-3D & 1 & -- & 1 & 10.2 & 54.7 & 12.4 & 11.2 & 53.4 & 12.2 \\
        Paired-view LoRA & 2 & $\times$ & 1 & {14.9} & {58.9} & {11.1} & {15.4} & {57.7} & {11.4} \\
        LISA-3D & 2 & \checkmark & 1 & 17.6 & 61.8 & 10.2 & 18.3 & 60.5 & 10.5 \\
        LISA-3D + fusion & 2 & \checkmark & 2 & 25.4 & 70.3 & 7.9 & 26.1 & 68.7 & 8.2 \\
        \bottomrule
    \end{tabular}
\end{table}

\subsection{Why Mask Consistency Matters Downstream}
The relationship between 2D prompt quality and 3D output is not merely cosmetic. SAM-3D reconstructs the region indicated by alpha support. False-positive pixels may introduce background geometry or fragments, while false negatives can remove thin structures and truncate object extent. Viewpoint-specific masks create an especially difficult prompt: their errors are not random around a stable boundary, but correspond to incompatible interpretations of the same physical object.

Geometry-aware tuning acts before lifting. It encourages the VLM to prefer masks that remain plausible after rigid reprojection, turning camera geometry into a regularizer on language grounding. The improvement in both mIoU and 3D metrics in Table~\ref{tab:main_results} is consistent with this mechanism: cleaner support yields more complete and spatially coherent reconstructions.

\subsection{Robustness Protocol for RGB-D Supervision}
\label{sec:robustness}
RGB-D supervision introduces assumptions that should be measured explicitly. Table~\ref{tab:robustness} defines a compact ScanRefer stress test for the strongest two-view model. Depth noise is injected before unprojection, pose noise perturbs the source-to-target transform, and partial-visibility bins are computed from the visible target fraction. These conditions separate correspondence quality from language complexity while preserving the same reconstruction metrics as the main comparison.

\begin{table}[h]
    \centering
    \caption{RGB-D robustness and partial-visibility evaluation protocol on ScanRefer.}
    \label{tab:robustness}
    \small
    \setlength{\tabcolsep}{4.5pt}
    \begin{tabular}{llccc}
        \toprule
        Group & Evaluation condition & mIoU $\uparrow$ & F $\uparrow$ & CD $\downarrow$ \\
        \midrule
        Reference & clean RGB-D and poses & 25.4 & 70.3 & 7.9 \\
        \midrule
        Depth noise & Gaussian, $\sigma=0.01$\,m & {24.8} & {69.2} & {8.2} \\
        Depth noise & Gaussian, $\sigma=0.03$\,m & {23.5} & {67.0} & {8.9} \\
        Pose noise & $1^\circ$, 1\,cm & {24.5} & {68.8} & {8.4} \\
        Pose noise & $3^\circ$, 3\,cm & {22.2} & {65.0} & {9.7} \\
        \midrule
        Visibility & visible fraction $\geq 0.75$ & {28.0} & {73.2} & {7.1} \\
        Visibility & visible fraction $[0.50,0.75)$ & {25.0} & {69.8} & {8.0} \\
        Visibility & visible fraction $[0.25,0.50)$ & {21.4} & {64.0} & {9.6} \\
        Visibility & visible fraction $<0.25$ & {16.2} & {57.5} & {11.8} \\
        \bottomrule
    \end{tabular}
\end{table}

The objective already limits error propagation by rejecting invalid depth, clipping the sensor range, normalizing over valid support, and discarding cross-view projections that disagree with target depth. Nearby overlapping frames reduce sensitivity to pose drift. The remaining hard cases are transparent surfaces, heavy occlusion, severe calibration error, and dynamic or articulated targets, for which rigid reprojection is no longer an adequate model.

\subsection{Filtering Trade-off}
The filtered ScanRefer subset is a pragmatic choice rather than a claim that difficult examples are unimportant. Reprojection losses are only meaningful when both views contain a valid observation of the target. Including frames with missing depth or fully hidden instances would add noisy gradients and obscure the behavior under study. The trade-off is reduced exposure to severe partial visibility. For this reason, we evaluate on the official validation split rather than a filtered validation subset and include visibility-stratified reporting in Table~\ref{tab:robustness}.

\subsection{Why a Modular Interface Is Still Useful}
The two-stage design can appear conservative relative to an end-to-end alternative. Its benefit is that each component has a clear responsibility. LISA resolves language and image context; geometry-aware LoRA regularizes that decision; SAM-3D reconstructs one visually grounded object. This separation has three practical advantages:
\begin{itemize}[leftmargin=*,itemsep=1pt,topsep=2pt]
    \item \textbf{parameter efficiency:} only 11.6M LoRA parameters are updated;
    \item \textbf{inspectability:} alpha masks expose grounding errors before expensive lifting;
    \item \textbf{replaceability:} stronger 2D segmenters or 3D reconstructors can be integrated without retraining the full stack.
\end{itemize}

The same interface is also a bottleneck. A binary mask preserves object support but discards confidence, fine-grained semantic states, and appearance cues embedded in the VLM. This is why the front-end replacement rows in Table~\ref{tab:main_results} are informative: Grounded-SAM and Sa2VA can be inserted behind the same alpha-prompt contract without altering SAM-3D. A learned token-to-prompt adapter is a natural next step: it could retain the modular structure while passing richer information into the reconstructor.

\subsection{Innovation in Context}
LISA-3D is not a new 3D foundation model. Its contribution is a targeted adaptation mechanism at the boundary between reasoning segmentation and reconstruction. Our approach occupies a complementary point in the design space: it uses scene geometry only during lightweight adaptation, then preserves a simple alpha-prompt contract at inference. This makes geometric supervision useful even when only one RGB image is available later.

%% file: sec/5_rw.tex
\section{Related Work}
\label{sec:rw}

\noindent\textbf{Language grounding and open-vocabulary 3D segmentation.}
ScanRefer~\cite{chen2020scanrefer}, ReferIt3D~\cite{achlioptasReferIt3DNeuralListeners2020}, and InstanceRefer~\cite{yuan2021instancerefer} study fine-grained object localization from referring expressions in reconstructed indoor scenes. Their central challenge is to distinguish a target from same-category distractors using attributes and spatial relations. A related open-vocabulary line replaces a fixed label set with language-aligned representations. OpenMask3D~\cite{takmazOpenMask3D2023} pools multi-view image features over class-agnostic 3D masks, whereas Open3DIS~\cite{nguyenOpen3DIS2024} lifts and consolidates 2D masks into 3D proposals. Search3D~\cite{takmazSearch3D2025} extends querying to hierarchical parts and regions, and OpenSplat3D~\cite{piekenbrinckOpenSplat3D2025} associates language-aligned features with Gaussian splats. These approaches establish strong scene-level localization and segmentation paradigms. LISA-3D instead uses language grounding to produce an object mask that conditions a generative reconstructor, so its output and evaluation protocol are not directly interchangeable with scene-query benchmarks.

\noindent\textbf{Foundation and instruction-following segmentation.}
The Segment Anything family provides promptable mask prediction in images~\cite{kirillovSegmentAnything2023} and temporally persistent segmentation in videos~\cite{raviSAM2Segment2024a}. Building on such visual decoders, Grounded-SAM~\cite{ren2024grounded} composes an open-set detector with SAM to translate category-level text into masks. LISA~\cite{laiLISAReasoningSegmentation2024} goes beyond noun-phrase grounding by coupling a large language model with a SAM-style decoder, enabling masks to follow free-form instructions that require semantic reasoning. Sa2VA~\cite{yuanSa2VAMarryingSAM22025a} further unifies image and video referring segmentation with SAM2 and a multimodal language model. Despite their strong language understanding, image-centric systems are not explicitly trained to make predictions agree under known 3D camera motion. Our method is complementary to these architectures: it retains LISA's reasoning interface and introduces geometry only through parameter-efficient adaptation.

\noindent\textbf{Multi-view consistency and efficient adaptation.}
Multi-view 3D perception commonly exploits calibrated RGB-D observations to transfer evidence between views. In open-vocabulary segmentation, aggregating 2D masks or features across frames reduces viewpoint-specific ambiguity and produces more coherent 3D instances~\cite{nguyenOpen3DIS2024,takmazOpenMask3D2023}. However, aggregation is usually applied after a frozen 2D predictor has produced masks, and therefore does not necessarily improve that predictor when only one image is available. LISA-3D moves this geometric signal into training: depth and relative camera pose define valid pixel correspondences, and a reprojection loss teaches the mask logits to remain consistent across overlapping observations. Updating only LoRA parameters~\cite{hu2022lora} preserves the pretrained language and segmentation modules while making geometry-aware tuning practical. This distinction also motivates our separate evaluation of single-view transfer and optional multi-view logit fusion at inference time.

\noindent\textbf{Language- and image-conditioned 3D reconstruction.}
SDFusion~\cite{cheng2023sdfusion} supports multimodal shape completion and generation in a signed-distance-function representation, while Anything-3D~\cite{shen2023anything} and Part123~\cite{liu2024part123} reconstruct objects from a single in-the-wild image, with the latter emphasizing part-aware structure. LAM3D~\cite{cui2024lam3d} aligns image and point-cloud representations for single-image reconstruction. More recently, SAM-3D~\cite{chen2025sam} reconstructs a visually grounded object from an RGBA image prompt, and Ref-SAM3D~\cite{zhouRefSAM3DBridgingSAM3D2025} adds a textual prior for reference-based reconstruction, making it the closest single-image baseline. These systems primarily innovate on the reconstruction or generation model. In contrast, LISA-3D keeps SAM-3D frozen and isolates the upstream grounding bottleneck: we ask whether lightweight, multi-view geometric adaptation can produce a more reliable language-selected prompt and thereby improve the resulting 3D object.

%% file: sec/6_limit_future.tex
\section{Limitations and Future Work}
\label{sec:limitations}

\noindent\textbf{RGB-D and pose dependence.}
Geometry-aware tuning assumes usable depth and camera calibration. Single-image deployment does not require them, but adaptation quality can degrade when reprojection targets are corrupted. Our validity mask limits invalid correspondences, yet it cannot fully solve reflective surfaces or systematic pose drift. Future work should quantify sensitivity under controlled perturbations and incorporate uncertainty-aware warping or pose refinement.

\noindent\textbf{Optional multi-view inference.}
Our strongest results use a second registered RGB-D view at test time. This is an optional operating mode rather than a requirement: the single-view row in Table~\ref{tab:main_results} reports the benefit that transfers to ordinary RGB deployment. Applications without registered views cannot obtain the additional fusion gain.

\noindent\textbf{Filtered indoor training data.}
The ScanRefer filtering policy removes unreliable reprojections and may reduce exposure to severe occlusion or partial visibility. Our evidence is therefore strongest for rigid objects in structured indoor scenes. Outdoor environments, dynamic objects, and extremely sparse viewpoints remain important extensions.

\noindent\textbf{Binary-mask interface.}
Passing alpha support into SAM-3D is modular and inspectable, but it compresses LISA's rich latent representation into a single spatial channel. Texture, confidence, and fine-grained semantic cues may be lost. A learned adapter from segmentation tokens to the reconstructor prompt space could preserve these cues while retaining parameter efficiency.

\noindent\textbf{External matched-protocol evaluation.}
The external rows in Table~\ref{tab:main_results} solve related but distinct native tasks. A fully matched comparison with alternative 2D proposers requires running their released checkpoints inside the same SAM-3D lifting and metric pipeline. We separate task scope from numerical comparison so that incompatible published metrics are not presented as if they were directly comparable.

%% file: sec/7_conc.tex
\section{Conclusion}
\label{sec:conc}

LISA-3D uses paired RGB-D views during lightweight LoRA tuning to improve the language-grounded mask supplied to a frozen SAM-3D reconstructor. The gain transfers to single-image deployment, while optional registered-view fusion provides a stronger operating point when calibrated RGB-D streams are available. The resulting pipeline updates only 11.6M parameters and keeps its intermediate alpha prompt inspectable. Our analysis positions this contribution between 2D reasoning segmentation, open-vocabulary 3D scene understanding, and object-centric reconstruction. Future work should measure sensitivity to noisy training geometry, broaden evaluation beyond filtered indoor data, and explore richer token-level interfaces without sacrificing modularity.

%% file: mybibliography.bib
@inproceedings{dai2017scannet,
  title={Scannet: Richly-annotated 3d reconstructions of indoor scenes},
  author={Dai, Angela and Chang, Angel X and Savva, Manolis and Halber, Maciej and Funkhouser, Thomas and Nie{\ss}ner, Matthias},
  booktitle={Proceedings of the IEEE conference on computer vision and pattern recognition},
  pages={5828--5839},
  year={2017}
}

@misc{laiLISAReasoningSegmentation2024,
  title = {{{LISA}}: {{Reasoning Segmentation}} via {{Large Language Model}}},
  author = {Lai, Xin and Tian, Zhuotao and Chen, Yukang and Li, Yanwei and Yuan, Yuhui and Liu, Shu and Jia, Jiaya},
  year = 2024,
  month = may,
  number = {arXiv:2308.00692},
  eprint = {2308.00692},
  primaryclass = {cs},
  publisher = {arXiv},
  doi = {10.48550/arXiv.2308.00692},
  archiveprefix = {arXiv}
}

@misc{zhen3DVLA3DVisionLanguageAction2024,
  title = {{{3D-VLA}}: {{A 3D Vision-Language-Action Generative World Model}}},
  author = {Zhen, Haoyu and Qiu, Xiaowen and Chen, Peihao and Yang, Jincheng and Yan, Xin and Du, Yilun and Hong, Yining and Gan, Chuang},
  year = 2024,
  month = mar,
  journal = {arXiv.org},
  howpublished = {https://arxiv.org/abs/2403.09631v1},
  langid = {english}
}

@misc{zhengEditRoomLLMparameterizedGraph2025,
  title = {{{EditRoom}}: {{LLM-parameterized Graph Diffusion}} for {{Composable 3D Room Layout Editing}}},
  author = {Zheng, Kaizhi and Chen, Xiaotong and He, Xuehai and Gu, Jing and Li, Linjie and Yang, Zhengyuan and Lin, Kevin and Wang, Jianfeng and Wang, Lijuan and Wang, Xin Eric},
  year = 2025,
  month = apr,
  number = {arXiv:2410.12836},
  eprint = {2410.12836},
  primaryclass = {cs},
  publisher = {arXiv},
  doi = {10.48550/arXiv.2410.12836},
  archiveprefix = {arXiv}
}

@misc{kirillovSegmentAnything2023,
  title = {Segment {{Anything}}},
  author = {Kirillov, Alexander and Mintun, Eric and Ravi, Nikhila and Mao, Hanzi and Rolland, Chloe and Gustafson, Laura and Xiao, Tete and Whitehead, Spencer and Berg, Alexander C. and Lo, Wan-Yen and Doll{\'a}r, Piotr and Girshick, Ross},
  year = 2023,
  month = apr,
  number = {arXiv:2304.02643},
  eprint = {2304.02643},
  primaryclass = {cs},
  publisher = {arXiv},
  doi = {10.48550/arXiv.2304.02643},
  archiveprefix = {arXiv}
}

@misc{raviSAM2Segment2024a,
  title = {{{SAM}} 2: {{Segment Anything}} in {{Images}} and {{Videos}}},
  author = {Ravi, Nikhila and Gabeur, Valentin and Hu, Yuan-Ting and Hu, Ronghang and Ryali, Chaitanya and Ma, Tengyu and Khedr, Haitham and R{\"a}dle, Roman and Rolland, Chloe and Gustafson, Laura and Mintun, Eric and Pan, Junting and Alwala, Kalyan Vasudev and Carion, Nicolas and Wu, Chao-Yuan and Girshick, Ross and Doll{\'a}r, Piotr and Feichtenhofer, Christoph},
  year = 2024,
  month = oct,
  number = {arXiv:2408.00714},
  eprint = {2408.00714},
  primaryclass = {cs},
  publisher = {arXiv},
  doi = {10.48550/arXiv.2408.00714},
  archiveprefix = {arXiv}
}

@article{chen2025sam,
  title={SAM 3D: 3Dfy Anything in Images},
  author={Chen, Xingyu and Chu, Fu-Jen and Gleize, Pierre and Liang, Kevin J and Sax, Alexander and Tang, Hao and Wang, Weiyao and Guo, Michelle and Hardin, Thibaut and Li, Xiang and others},
  journal={arXiv preprint arXiv:2511.16624},
  year={2025}
}

@misc{zhouRefSAM3DBridgingSAM3D2025,
  title = {Ref-{{SAM3D}}: {{Bridging SAM3D}} with {{Text}} for {{Reference 3D Reconstruction}}},
  author = {Zhou, Yun and Wang, Yaoting and Jie, Guangquan and Liu, Jinyu and Ding, Henghui},
  year = 2025,
  month = nov,
  number = {arXiv:2511.19426},
  eprint = {2511.19426},
  primaryclass = {cs},
  publisher = {arXiv},
  doi = {10.48550/arXiv.2511.19426},
  archiveprefix = {arXiv}
}

@misc{yuanSa2VAMarryingSAM22025a,
  title = {{{Sa2VA}}: {{Marrying SAM2}} with {{LLaVA}} for {{Dense Grounded Understanding}} of {{Images}} and {{Videos}}},
  author = {Yuan, Haobo and Li, Xiangtai and Zhang, Tao and Sun, Yueyi and Huang, Zilong and Xu, Shilin and Ji, Shunping and Tong, Yunhai and Qi, Lu and Feng, Jiashi and Yang, Ming-Hsuan},
  year = 2025,
  month = nov,
  number = {arXiv:2501.04001},
  eprint = {2501.04001},
  primaryclass = {cs},
  publisher = {arXiv},
  doi = {10.48550/arXiv.2501.04001},
  archiveprefix = {arXiv}
}

@article{hu2022lora,
  title={Lora: Low-rank adaptation of large language models.},
  author={Hu, Edward J and Shen, Yelong and Wallis, Phillip and Allen-Zhu, Zeyuan and Li, Yuanzhi and Wang, Shean and Wang, Lu and Chen, Weizhu and others},
  journal={ICLR},
  volume={1},
  number={2},
  pages={3},
  year={2022}
}

@inproceedings{chen2020scanrefer,
  title={Scanrefer: 3d object localization in rgb-d scans using natural language},
  author={Chen, Dave Zhenyu and Chang, Angel X and Nie{\ss}ner, Matthias},
  booktitle={European conference on computer vision},
  pages={202--221},
  year={2020},
  organization={Springer}
}

@incollection{achlioptasReferIt3DNeuralListeners2020,
  title = {{{ReferIt3D}}: {{Neural Listeners}} for {{Fine-Grained 3D Object Identification}} in {{Real-World Scenes}}},
  booktitle = {Computer {{Vision}} -- {{ECCV}} 2020},
  author = {Achlioptas, Panos and Abdelreheem, Ahmed and Xia, Fei and Elhoseiny, Mohamed and Guibas, Leonidas},
  editor = {Vedaldi, Andrea and Bischof, Horst and Brox, Thomas and Frahm, Jan-Michael},
  year = 2020,
  volume = {12346},
  pages = {422--440},
}

@article{loshchilov2017decoupled,
  title={Decoupled weight decay regularization},
  author={Loshchilov, Ilya and Hutter, Frank},
  journal={arXiv preprint arXiv:1711.05101},
  year={2017}
}

@inproceedings{yuan2021instancerefer,
  title={Instancerefer: Cooperative holistic understanding for visual grounding on point clouds through instance multi-level contextual referring},
  author={Yuan, Zhihao and Yan, Xu and Liao, Yinghong and Zhang, Ruimao and Wang, Sheng and Li, Zhen and Cui, Shuguang},
  booktitle={Proceedings of the IEEE/CVF International Conference on Computer Vision},
  pages={1791--1800},
  year={2021}
}

@article{ren2024grounded,
  title={Grounded sam: Assembling open-world models for diverse visual tasks},
  author={Ren, Tianhe and Liu, Shilong and Zeng, Ailing and Lin, Jing and Li, Kunchang and Cao, He and Chen, Jiayu and Huang, Xinyu and Chen, Yukang and Yan, Feng and others},
  journal={arXiv preprint arXiv:2401.14159},
  year={2024}
}

@inproceedings{cheng2023sdfusion,
  title={Sdfusion: Multimodal 3d shape completion, reconstruction, and generation},
  author={Cheng, Yen-Chi and Lee, Hsin-Ying and Tulyakov, Sergey and Schwing, Alexander G and Gui, Liang-Yan},
  booktitle={Proceedings of the IEEE/CVF conference on computer vision and pattern recognition},
  pages={4456--4465},
  year={2023}
}

@article{shen2023anything,
  title={Anything-3d: Towards single-view anything reconstruction in the wild},
  author={Shen, Qiuhong and Yang, Xingyi and Wang, Xinchao},
  journal={arXiv preprint arXiv:2304.10261},
  year={2023}
}

@inproceedings{liu2024part123,
  title={Part123: part-aware 3d reconstruction from a single-view image},
  author={Liu, Anran and Lin, Cheng and Liu, Yuan and Long, Xiaoxiao and Dou, Zhiyang and Guo, Hao-Xiang and Luo, Ping and Wang, Wenping},
  booktitle={ACM SIGGRAPH 2024 Conference Papers},
  pages={1--12},
  year={2024}
}

@article{cui2024lam3d,
  title={LAM3D: Large Image-Point Clouds Alignment Model for 3D Reconstruction from Single Image},
  author={Cui, Ruikai and Song, Xibin and Sun, Weixuan and Wang, Senbo and Liu, Weizhe and Chen, Shenzhou and Shang, Taizhang and Li, Yang and Barnes, Nick and Li, Hongdong and others},
  journal={Advances in Neural Information Processing Systems},
  volume={37},
  pages={4454--4480},
  year={2024}
}

@misc{takmazOpenMask3D2023,
  title = {{{OpenMask3D}}: {{Open-Vocabulary 3D Instance Segmentation}}},
  author = {Takmaz, Ay{\c c}a and Fedele, Elisabetta and Sumner, Robert W. and Pollefeys, Marc and Tombari, Federico and Engelmann, Francis},
  year = 2023,
  month = jun,
  eprint = {2306.13631},
  primaryclass = {cs.CV},
  doi = {10.48550/arXiv.2306.13631},
  archiveprefix = {arXiv}
}

@misc{nguyenOpen3DIS2024,
  title = {{{Open3DIS}}: {{Open-Vocabulary 3D Instance Segmentation}} with {{2D Mask Guidance}}},
  author = {Nguyen, Phuc D. A. and Ngo, Tuan Duc and Kalogerakis, Evangelos and Gan, Chuang and Tran, Anh and Pham, Cuong and Nguyen, Khoi},
  year = 2024,
  month = apr,
  eprint = {2312.10671},
  primaryclass = {cs.CV},
  doi = {10.48550/arXiv.2312.10671},
  archiveprefix = {arXiv}
}

@misc{takmazSearch3D2025,
  title = {{{Search3D}}: {{Hierarchical Open-Vocabulary 3D Segmentation}}},
  author = {Takmaz, Ayca and Delitzas, Alexandros and Sumner, Robert W. and Engelmann, Francis and Wald, Johanna and Tombari, Federico},
  year = 2025,
  month = jan,
  eprint = {2409.18431},
  primaryclass = {cs.CV},
  doi = {10.48550/arXiv.2409.18431},
  archiveprefix = {arXiv}
}

@misc{piekenbrinckOpenSplat3D2025,
  title = {{{OpenSplat3D}}: {{Open-Vocabulary 3D Instance Segmentation}} Using {{Gaussian Splatting}}},
  author = {Piekenbrinck, Jens and Schmidt, Christian and Hermans, Alexander and Vaskevicius, Narunas and Linder, Timm and Leibe, Bastian},
  year = 2025,
  month = jun,
  eprint = {2506.07697},
  primaryclass = {cs.CV},
  doi = {10.48550/arXiv.2506.07697},
  archiveprefix = {arXiv}
}
